\documentclass[conference]{IEEEtran}

\usepackage{cite}
\usepackage{amsmath,amssymb,amsfonts}
\usepackage{algorithmic}
\usepackage{graphicx}
\usepackage{textcomp}
\usepackage{wrapfig}
\usepackage{balance}
\usepackage{multirow}
\usepackage{stmaryrd}
\usepackage[hidelinks]{hyperref}
\usepackage{tabularx}
\usepackage{makecell}

\def\BibTeX{{\rm B\kern-.05em{\sc i\kern-.025em b}\kern-.08em
    T\kern-.1667em\lower.7ex\hbox{E}\kern-.125emX}}

\begin{document}

\title{A dataset for audio-video based vehicle speed estimation}

\author{
	\IEEEauthorblockN{1\textsuperscript{st} Slobodan Djukanovi\'c}
	\IEEEauthorblockA{\textit{Faculty of Electrical Engineering} \\
		\textit{University of Montenegro}\\
		Podgorica, Montenegro \\
		slobdj@ucg.ac.me}
	\and
	\IEEEauthorblockN{2\textsuperscript{nd} Nikola Bulatovi\'c}
	\IEEEauthorblockA{\textit{Faculty of Electrical Engineering} \\
		\textit{University of Montenegro}\\
		Podgorica, Montenegro \\
		nbulatovic@ucg.ac.me}
	\and
	\IEEEauthorblockN{3\textsuperscript{rd} Ivana \v{C}avor}
	\IEEEauthorblockA{\textit{Faculty of Maritime Studies} \\
		\textit{University of Montenegro}\\
		Kotor, Montenegro \\
		ivana.ca@ucg.ac.me}
}
\maketitle

\begin{abstract}
Accurate speed estimation of road vehicles is important for several reasons. One is speed limit enforcement, which represents a crucial tool in decreasing traffic accidents and fatalities. Compared with other research areas and domains, the number of available datasets for vehicle speed estimation is still very limited. We present a dataset of on-road audio-video recordings of single vehicles passing by a camera at known speeds, maintained stable by the on-board cruise control. The dataset contains thirteen vehicles, selected to be as diverse as possible in terms of manufacturer, production year, engine type, power and transmission, resulting in a total of $ 400 $ annotated audio-video recordings. The dataset is fully available and intended as a public benchmark to facilitate research in audio-video vehicle speed estimation. In addition to the dataset, we propose a cross-validation strategy which can be used in a machine learning model for vehicle speed estimation. Two approaches to training-validation split of the dataset are proposed.
\end{abstract}


\section{Introduction}
\label{Intro}
The demands of improving traffic safety, mobility and efficiency, reducing air pollution and mitigating the impact of traffic problems (e.g. the impact of congestion on the economy) have led to the introduction of Intelligent Transport System (ITS). Therefore, an ITS applies a combination of leading-edge information and communication technologies to carry out sensing, analysis, control and communication tasks. 

One of the key ITS features is the ability of accurate speed estimation of the road vehicles, which is important for various reasons. First, since speeding increases both the risk of traffic accidents and the severity of consequences \cite{world2008speed}, speed limit enforcement is considered a crucial tool in decreasing traffic accidents and fatalities. For example, in the vicinity of speed cameras, the number of speeding vehicles and crashes is reduced up to $ 35 $\% and $ 25 $\%, respectively. In addition, the rate of accident reduction is directly proportional to the intensity or level of enforcement \cite{elvik2011developing}. Consequently, the number of speed cameras installed worldwide has been constantly growing. Second important reason is that the knowledge of the traffic speed (average speed of all vehicles in multiple road segments) can be used in adaptive traffic signal control, real-time traffic aware navigation, detection of traffic jams and accidents etc.

Speed estimation of the road vehicles encompasses multiple tasks, including synchronized data recording, representation, detection and tracking, as well as distance and speed estimation. The data are collected by sensors of different nature, such as radars, lasers or cameras. Depending on the application, the accuracy and robustness of speed estimation can be at different levels. Regarding speed enforcement, the estimation accuracy has to be very high, since speed offenders can be fined, lose their driver's license or even get imprisoned as a consequence. Therefore, sensors commonly used for speed measurement in sensitive applications are high-precision range sensors, such as radar (which uses the Doppler effect), LiDAR (based on the time of flight), or intrusive sensors embedded in pairs under the road surface (piezoelectric sensors and induction loops). These sensors provide very accurate speed estimations, although high price and costs of installation and maintenance impede their widespread use.

The focus of this paper are datasets for vehicle speed estimation. We consider video and audio data, i.e. data obtained from video cameras. The available audio-video datasets are described in Section \ref{AvailableDatasets}. Section \ref{VS13dataset} describes a dataset of $ 400 $ audio-video recordings collected for the purpose of vehicle speed estimation. Section \ref{Conclusion} concludes the paper.

\section{Available datasets}\label{AvailableDatasets}
\subsection{Video datasets}
In traffic monitoring, vision offers several important advantages with respect to other modalities. Cameras provide rich vehicle-based information such as visual features of vehicles, their geometry and path. A single camera can be utilized to detect and classify vehicles in multiple lanes. In addition, installation and maintenance of cameras in roadways is significantly less expensive and less disruptive than in intrusive systems. Even though vehicle speed estimation is a popular research topic, the number of available datasets is still very limited compared to other research areas. 

Recently, the most widely used dataset for traffic and vehicle speed detection has been the AI City Challenge \cite{naphade21aic21}. It is part of an annual challenge, originally proposed by NVIDIA in 2017, with the aim of pushing the boundaries of research and development in intelligent video analysis for various use cases in smart cities. Each year, the number of videos and samples provided varies. Specifically, the challenge in $ 2018 $ focused on problems such as estimating traffic flow characteristics, including traffic speed \cite{naphade2018aic18}. For the challenge purpose, $ 142 $ videos of different resolutions and lengths were provided, with no ground truth value of speed.

The BrnoCompSpeed dataset \cite{sochor2018comprehensive} contains $ 21 $ full-HD ($ 1920\times1080 $ pixels) videos, each with length of approximately $ 1 $ hour. Vehicles in the videos ($ 20 865 $ in total) were annotated using LiDAR and verified with several reference GPS tracks. 

Another useful dataset has been collected by the Federal University of Technology of Paraná \cite{luvizon2016video}. It contains $ 20 $ full-HD videos of total time of $ 291 $ minutes. The videos, containing $ 8849 $ vehicles in total, were divided into five sets depending on the weather and recording conditions. The ground truth speeds were obtained using a high precision speed meter based on an inductive loop detector. The bulk of the recorded speeds in the dataset is within range $ 40-60 $ km/h.

A thorough analysis of the vision-based vehicle speed estimation topic is given in \cite{fernandez2021vision}.

\subsection{Audio datasets}
Vision-based vehicle speed estimation suffers from reduced performance due to partial occlusion, shadows and illumination variation. What is more, video processing can be computationally expensive and time consuming. In terms of traffic monitoring, acoustic sensors offer many advantages over cameras. The list includes lower price, lower energy consumption, significantly reduced storage space, lower installation and maintenance costs \cite{won2020intelligent}.

Similarly to vision-based vehicle speed estimation, research progress in acoustic vehicle speed estimation is limited by the lack of rich datasets, preferably annotated ones. Datasets used in experiments in the studies \cite{cevher2008vehicle,barnwal2013doppler,lopez2004estimation,giraldo2016vehicle,koops2015ensemble,marmaroli2013pass,goksu2018vehicle} are usually very small. For example, Cevher et al. use ten audio recordings, corresponding to nine different vehicles \cite{cevher2008vehicle}. In \cite{barnwal2013doppler} and \cite{lopez2004estimation}, seven recordings were used, corresponding to three cars, a bus and a motorbike. Two different cars, four different speeds per car and two recordings per speed, were used in \cite{giraldo2016vehicle}. Research \cite{koops2015ensemble} used the sound of a car driven on a parking lot, with no car manufacturer, speed or the number of laps specified. One recording, with the length of $ 240 $ seconds, containing $ 22 $ cars and $ 2 $ motorbikes, was used in \cite{marmaroli2013pass}. G{\"o}ksu \cite{goksu2018vehicle} used the sound of one vehicle, with speeds ranging from $ 30 $ km/h to $ 80 $ km/h.

\section{VS13 dataset}\label{VS13dataset}
This paper presents a dataset of on-road audio-video recordings of vehicles passing by the camera at constant speeds. Each recording in the dataset contains a single drive of a single vehicle. Thirteen different vehicles were used, with a total of $ 400 $ annotated audio-video recordings. The dataset, referred to as VS$13$, is fully available, intended as a public benchmark to facilitate research on audio-video vehicle speed estimation.

\begin{figure}[t!]
	\centering
	\includegraphics{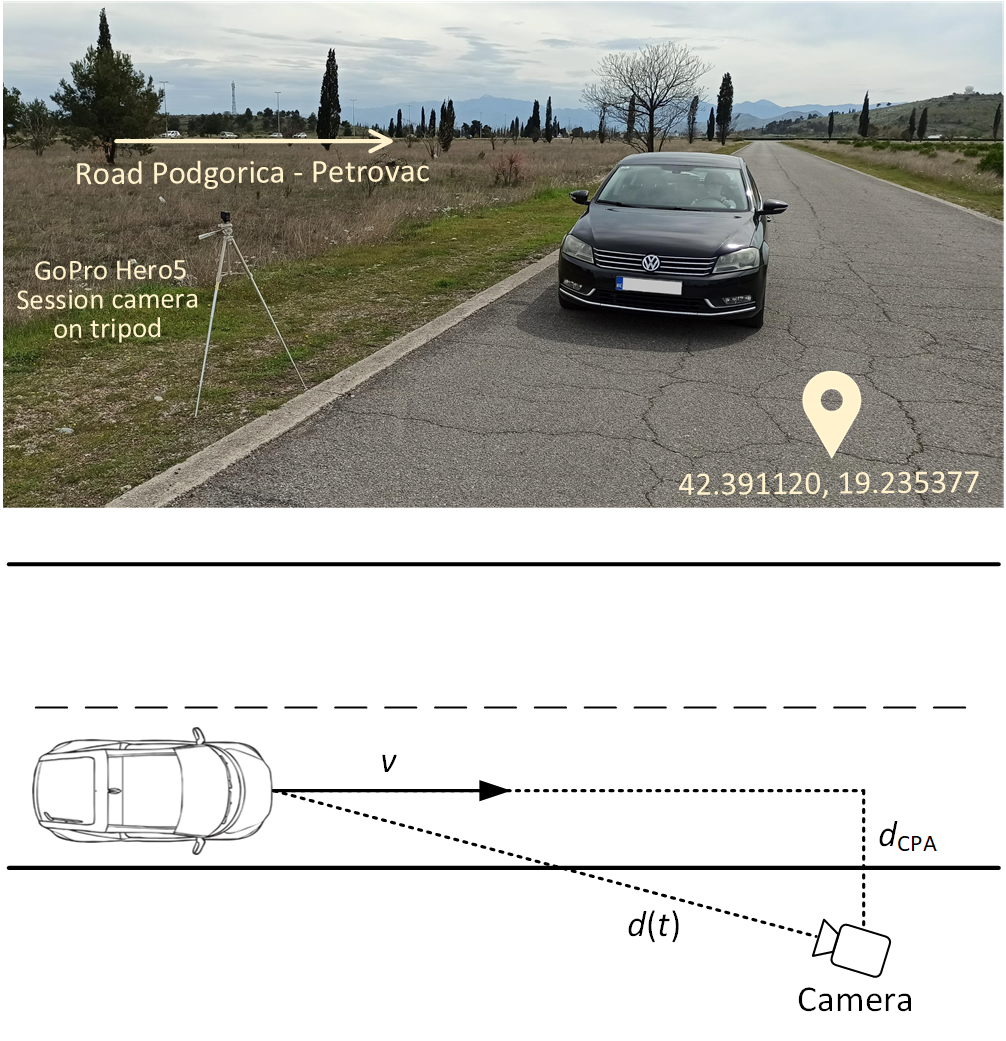}
	\caption{{\it{Top:}} Screenshot of the recording setup. GoPro Hero5 Session camera mounted on a tripod, at a distance of $ \approx 0.5 $ m from the road and at a height of $ \approx 1.2 $ m. {\it{Bottom:}} Vehicle moving at a constant speed $ v $. $ d(t) $ is the distance between a vehicle and the camera at time instant $ t $, whereas $ d_{\text{CPA}} $ is the distance at the closest point of approach (CPA).}
	\label{Fig1}
\end{figure}

VS$13$ has been compiled following these requirements \cite{djukanovic2021acoustic}:
\begin{enumerate}
	\item each recording contains a single drive of a single vehicle,
	\item recordings are made in an urban environment,
	\item recordings are real field ones,
	\item vehicles are as diverse as possible in terms of manufacturer, production year, engine type (petrol or diesel), power and transmission (manual or automatic),
	\item vehicles are equipped with cruise control (speed control), so that speed can be maintained stable while the vehicle passes by the camera.
\end{enumerate}

The first requirement implies that we address estimating the speed of individual vehicles rather than measuring the average speed of all vehicles along a road segment, i.e. the traffic speed. Requirements 2) and 3) are important for the context of acoustic vehicle speed estimation. More precisely, the two imply that the pass-by sound of vehicles (prominent sound source) may be corrupted by the sound of other nearby vehicles and natural sounds (e.g. wind, bird chirps, crickets).

The dataset (recordings with annotations) is available for download at \url{http://slobodan.ucg.ac.me/science/vs13/}. Audio files have been extracted from the recordings and provided separately for download, to facilitate research in acoustic vehicle speed estimation.

\subsection{Dataset collection}

The dataset was recorded on a local road, $ 622 $ m long, located $ 90 $ m away from the main road connecting the cities of Podgorica and Petrovac in Montenegro (see arrow in Fig. \ref{Fig1} (top)). The reasons for selecting this road are as follows \cite{djukanovic2021acoustic}:
\begin{itemize}
	\item it is long enough so that stable speeds can be achieved prior to the pass-by instant,
	\item it is isolated enough to allow measurements without too many disturbances,
	\item it is close to other roads so that requirements 2) and 3) are met.
\end{itemize}

\begin{table*}[t!]
	\centering
	\caption{VS$ 13 $ vehicles and speeds}
	\begin{tabularx}{\textwidth}{ l | c | c | c | c | c | c }
		\hline\hline
		\multicolumn{1}{c|}{\makecell[c]{Vehicle\\(Short name)}} & \multicolumn{1}{c|}{\makecell[c]{Engine\\type}} & \multicolumn{1}{c|}{\makecell[c]{Power\\(kW)}} & \multicolumn{1}{c|}{Transmission} & \multicolumn{1}{c|}{\makecell[c]{Prod.\\year}} & \multicolumn{1}{c|}{\makecell[c]{Record.\\sessions}} & \multicolumn{1}{c}{Speeds (km/h)} \\ \hline
		\makecell[l]{Citroen C4 Picasso 1.6 HDI\\(CitroenC4Picasso)} & Diesel & 88 & Manual & 2015 & 1 & \makecell[l]{35, 38, 41, 44, 48, 51, 54, 57, 59, 63, 65, 68, 72, 74, 78, 80, 83, 85, 87, \\92, 94, 96, 101} \\ \hline
		\makecell[l]{Kia Sportage 1.6 GDI\\(KiaSportage)} & Petrol & 97 & Manual & 2021 & 1 & \makecell[l]{31, 33, 35, 38, 41, 44, 46, 48, 51, 53, 55, 58, 61, 63, 65, 68, 69, 72, 74, \\77, 78, 80, 83, 85, 86, 89, 91, 93, 96, 98, 100, 103, 105} \\ \hline
		\makecell[l]{Mazda 3 Skyactive\\(Mazda3)} & Petrol & 74 & Manual & 2015 & 1 & \makecell[l]{30, 33, 35, 38, 40, 43, 45, 47, 50, 52, 55, 57, 60, 62, 64, 67, 70, 72, 75, \\79, 81, 84, 86, 88, 90, 92, 94, 96, 99, 101, 103, 105} \\ \hline
		\makecell[l]{Mercedes AMG 550\\(MercedesAMG550)} & Petrol & 350 & Automatic & 2006 & 3 & \makecell[l]{30, 33, 35, 38, 40, 42, 45, 47, 50, 52, 55, 58, 60, 62, 65, 67, 70, 73, 75, \\78, 80, 82, 85, 87, 90, 93, 95, 98, 100, 105} \\ \hline
		\makecell[l]{Mercedes GLA 200D\\(MercedesGLA)} & Diesel & 100 & Automatic & 2017 & 1 & \makecell[l]{30, 33, 36, 39, 41, 42, 45, 47, 48, 49, 52, 54, 55, 59, 61, 63, 65, 68, 70, \\72, 75, 78, 81, 83, 85, 88, 90, 92, 93, 96, 100, 101, 103, 104} \\ \hline
		\makecell[l]{Nissan Qashqai 1.5 DCI\\(NissanQashqai)} & Diesel & 81 & Manual & 2018 & 1 & \makecell[l]{35, 38, 40, 42, 45, 48, 50, 53, 55, 58, 60, 61, 64, 65, 68, 70, 73, 75, 78, \\80, 82, 85, 88, 90, 93, 94, 96, 98, 102} \\ \hline
		\makecell[l]{Opel Insignia 2.0 CDTI\\(OpelInsignia)} & Diesel & 96 & Automatic & 2010 & 1 & \makecell[l]{31, 35, 38, 41, 44, 47, 50, 53, 55, 58, 61, 64, 66, 68, 70, 72, 73, 76, 78, \\80, 83, 86, 89, 91, 94, 97, 100} \\ \hline
		\makecell[l]{Peugeot 208 1.4 HDI\\(Peugeot208)} & Diesel & 50 & Automatic & 2014 & 1 & \makecell[l]{30, 32, 34, 37, 40, 43, 45, 47, 50, 51, 54, 57, 60, 62, 64, 67, 68, 71, 73, \\76, 77, 79, 82, 84, 87, 90, 92, 95, 96} \\ \hline
		\makecell[l]{Peugeot 3008 1.6 HDI\\(Peugeot3008)} & Diesel & 84 & Automatic & 2013 & 2 & \makecell[l]{40, 43, 45, 47, 50, 52, 54, 55, 56, 58, 60, 61, 63, 65, 67, 68, 70, 72, 74, \\75, 78, 80, 83, 85, 87, 89, 90, 92, 95, 97, 100} \\ \hline
		\makecell[l]{Peugeot 307 2.0 HDI\\(Peugeot307)} & Diesel & 100 & Manual & 2007 & 1 & \makecell[l]{30, 33, 35, 38, 40, 43, 45, 47, 48, 50, 53, 56, 59, 60, 63, 66, 69, 72, 73, \\76, 79, 82, 85, 88, 91, 94, 97, 101, 103} \\ \hline
		\makecell[l]{Renault Captur 1.5 DCI\\(RenaultCaptur)} & Diesel & 66 & Automatic & 2015 & 1 & \makecell[l]{30, 33, 36, 38, 40, 41, 44, 46, 47, 48, 50, 52, 56, 58, 60, 63, 66, 68, 70, \\72, 76, 78, 80, 83, 86, 88, 90, 92, 94, 97, 98, 100, 102} \\ \hline
		\makecell[l]{Renault Scenic 1.9 DCI\\(RenaultScenic)} & Diesel & 96 & Manual & 2010 & 2 & \makecell[l]{30, 35, 36, 38, 40, 42, 44, 46, 48, 50, 52, 54, 57, 60, 62, 64, 66, 68, 70, \\71, 72, 74, 75, 77, 80, 82, 84, 86, 87, 90, 91, 94, 95, 98, 101} \\ \hline
		\makecell[l]{VW Passat B7 1.6 TDI\\(VWPassat)} & Diesel & 77 & Manual & 2011 & 2 & \makecell[l]{30, 35, 39, 40, 42, 45, 47, 49, 50, 52, 54, 55, 57, 60, 61, 64, 65, 67, 70, \\71, 72, 73, 75, 78, 80, 81, 82, 85, 88, 90, 91, 94, 96, 98, 100} \\ \hline	
		\hline
	\end{tabularx}
	\label{Tab1}
\end{table*}

A GoPro Hero5 Session camera was used for recording the dataset. It was installed by the road, mounted on a tripod, approximately at $ 0.5 $ m from the road and at $ 1.2 $ m height. Screenshot of the recording setup is presented in Fig. \ref{Fig1} (top). The camera position varied with respect to the road, that is, it was installed on both sides of the road and at different angles\footnote{Sample video files of each vehicle can be seen at \url{http://slobodan.ucg.ac.me/science/vs13/}.}. The recording sessions (one session per day) took place from December 2019 to February 2022. Thirteen vehicles were used, as listed in Table \ref{Tab1}. The number of recording sessions per vehicle is given in the sixth column of Table \ref{Tab1}.

\subsection{Dataset speeds}
\begin{figure}[t!]
	\centering
	\includegraphics{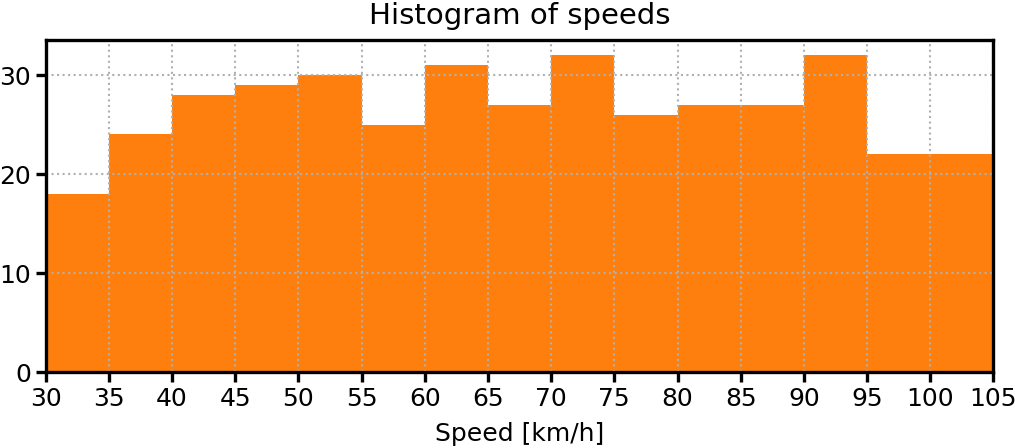}
	\caption{Histogram of VS$ 13 $ speeds ($ 15 $ equal-width bins).}
	\label{Fig2}
\end{figure}

Speeds in the dataset range from $30$ to $105$ km/h, with the exact values given in the last column in Table \ref{Tab1}. For lower speeds, under $ 30 $ km/h, the cruise control cannot operate with the selected vehicles (for Peugeot 3008, even below $ 40 $ km/h). For higher speeds, above $ 105 $ km/h, we couldn't carry out stable and secure measurements on the selected road. The speed step varies between $ 1 $ to $ 3 $ km/h and all speeds from $ 30 $ to $ 105 $ km/h are included in VS$ 13 $. Histogram of speeds is given in Fig. \ref{Fig2}. The reported speeds are stable at least $ 3 $ seconds before and after the pass by. Outside that $ 6 $-second interval, minor speed variations are possible.

\subsection{Dataset preprocessing: Video and audio}
The original recordings were cut into $ 10 $-second video files (MP4 format, full HD, $ 30 $ fps) so that the pass-by instants of vehicles are around the middle of the file. For that purpose, we used the Format Factory application.

VS$ 13 $ contains $ 13 $ folders, with $ 400 $ video files in total. Each folder contains $ 10 $-second video files and text annotations that correspond to one vehicle.

For the purpose of acoustic vehicle speed estimation, we extracted audio files ($ 44100 $ Hz sampling rate, WAV format, $ 32 $-bit float PCM) from the corresponding video files using the Audacity application. The extracted audio files and the corresponding annotations are also available for download.

\subsection{Dataset annotations}
Video and audio files in VS$ 13 $ are accompanied by annotation text files which contain two pieces of data each: the vehicle's speed and the pass-by-camera instant (with two-decimal precision). Relative time from the beginning of the file is given, measured in seconds. The pass-by-camera instant corresponds to the timestamp of a video frame that contains the vehicle starting to exit the camera view. That instant approximately corresponds to the closest point of approach (CPA), as depicted in Fig. \ref{Fig1} (bottom).

\subsection{Naming convention}
While naming the VS$ 13 $ files, the following convention has been respected:\\
\centerline{\texttt{shortVehicleName\_vehicleSpeed}}\\
where \texttt{shortVehicleName} and \texttt{vehicleSpeed} represent the short vehicle name (see the first column in Table \ref{Tab1}) and vehicle speed, respectively. For example, \texttt{MercedesGLA\_68.mp4}, \texttt{MercedesGLA\_68.wav} and \texttt{MercedesGLA\_68.txt} represent the names of video, audio and annotation files, respectively, of Mercedes GLA 200D driven at $ 68 $ km/h.

\subsection{Cross-validation strategy}\label{TrainValSplit}
In addition to the VS$ 13 $ dataset, we propose a cross-validation (CV) strategy which can be used in a machine learning model for vehicle speed estimation. Since there are $ 13 $ vehicles in the dataset, a natural solution would be to use $ 13 $-fold CV. One CV round (out of $ 13 $) implies that one vehicle is used for testing, whereas the remaining twelve ones are used for training and validating the model. For training-validation split, we propose two approaches. In the first approach, for each vehicle, the files are divided into the training and validation ones according to the $ 80 \%$--$ 20 \%$ rule. To that end, we carry out the following procedure:
\begin{enumerate}
	\item sort the speeds into ascending order,
	\item divide the sorted speeds into batches of five speeds,
	\item randomly select one speed in each batch to be used for validation and the other ones for training.
\end{enumerate}
This approach ensures that low-, medium- and high-speed files are used in both training and validation. Each VS$ 13 $ folder contains a file \texttt{Train\_valid\_split.txt} with labels \textit{train} or \textit{valid} associated with each file. Acoustic vehicle speed estimation methods \cite{djukanovic2021acoustic,bulatovic2022approach,bulatovic2022mel} use this strategy.

The second approach is to split the vehicles into the training and validation ones. For example, nine vehicles can be used for training and the remaining three for validation.

In order to evaluate the model as precisely as possible, the proposed two approaches can be implemented as iterated $ k $-fold CV with shuffling, which consists of applying $ k $-fold CV multiple times, shuffling the data every time before splitting it $ k $ ways. The final score is the average of the scores obtained at each run of $ k $-fold CV. Depending on the model architecture, this approach may require very high computational power.

\section{Conclusion}
\label{Conclusion}
In this paper, we presented a dataset of on-road audio-video recordings of vehicles passing by a camera at known constant speeds. The dataset, referred to as VS$ 13 $, contains $ 400 $ annotated audio-video recordings of vehicles, selected to be as diverse as possible in terms of manufacturer, production year, engine type, power and transmission. The dataset is publicly available, freely accessible, and intended as a public benchmark to facilitate research in vehicle speed estimation. In addition to the dataset, we proposed a cross-validation strategy which can be used in a machine learning model for vehicle speed estimation, as well as two approaches for training-validation dataset split.

\bibliographystyle{IEEEtran}
\bibliography{Bibliography}

\end{document}